\documentclass[10pt,journal,cspaper,compsoc]{IEEEtran}
\usepackage{graphicx}
\usepackage{amssymb}
\usepackage{amsmath}
\usepackage{multirow}
\usepackage{tabularx}
\usepackage{dsfont}

\usepackage{times}
\usepackage{epsfig}
\usepackage{graphicx}
\usepackage{amssymb}
\usepackage[pagebackref=true,breaklinks=true,letterpaper=true,colorlinks,bookmarks=false]{hyperref}
\usepackage{mdwmath}
\usepackage{hyperref}
\usepackage{array}
\usepackage{url}
\usepackage{times}
\usepackage{algorithm}
\usepackage{algorithmic}
\usepackage{subfigure}

\begin{document}
\title{CMS-RCNN: Contextual Multi-Scale Region-based CNN for Unconstrained Face Detection}

 \author{Chenchen~Zhu*, ~\IEEEmembership{Student,~IEEE,}%
         ~Yutong~Zheng*, ~\IEEEmembership{Student,~IEEE,}\\%
         ~Khoa~Luu, ~\IEEEmembership{Member,~IEEE,}%
%         ~T.~Hoang~Ngan~Le, ~\IEEEmembership{Student,~IEEE,}%
         ~Marios~Savvides, ~\IEEEmembership{Senior Member,~IEEE}\\%
 \thanks{CyLab Biometrics Center and the Department of Electrical and Computer Engineering, Carnegie Mellon University, Pittsburgh, PA, USA. Emails: \{chenchez, yutongzh, kluu\}@andrew.cmu.edu, msavvid@ri.cmu.edu
 
* indicates equal contribution. }}

\IEEEcompsoctitleabstractindextext{%
\begin{abstract}
Robust face detection in the wild is one of the ultimate components to support various facial related problems, i.e. unconstrained face recognition, facial periocular recognition, facial landmarking and pose estimation, facial expression recognition, 3D facial model construction, etc. Although the face detection problem has been intensely studied for decades with various commercial applications, it still meets problems in some real-world scenarios due to numerous challenges, e.g. heavy facial occlusions, extremely low resolutions, strong illumination, exceptionally pose variations, image or video compression artifacts, etc.
In this paper, we present a face detection approach named Contextual Multi-Scale Region-based Convolution Neural Network (CMS-RCNN) to robustly solve the problems mentioned above. Similar to the region-based CNNs, our proposed network consists of the region proposal component and the region-of-interest (RoI) detection component. However, far apart of that network, there are two main contributions in our proposed network that play a significant role to achieve the state-of-the-art performance in face detection. Firstly, the multi-scale information is grouped both in region proposal and RoI detection to deal with tiny face regions. Secondly, our proposed network allows explicit body contextual reasoning in the network inspired from the intuition of human vision system.
%uncooperative and non-cooperative
% The proposed approach is benchmarked on the challenging Wider Face dataset, and compared against two strong baselines, i.e. Pittpatt and Faster R-CNN. Experimental results show that our method consistently outperform the baselines by a large margin.
% Robust face detection is one of the most important pre-processing steps to support facial expression analysis, facial landmarking, face recognition, pose estimation, building of 3D facial models, etc.
% Although this topic has been intensely studied for decades, it is still challenging due to numerous variants of face images in real-world scenarios.
% In this paper, we present a novel approach named Multiple Scale Faster Region-based Convolutional Neural Network (MS-FRCNN) to robustly detect human facial regions from images collected under various challenging conditions, e.g. large occlusions, extremely low resolutions, facial expressions, strong illumination variations, etc.
The proposed approach is benchmarked on two recent challenging face detection databases, i.e. the WIDER FACE Dataset which contains high degree of variability, as well as the Face Detection Dataset and Benchmark (FDDB). The experimental results show that our proposed approach trained on WIDER FACE Dataset outperforms strong baselines on WIDER FACE Dataset by a large margin, and consistently achieves competitive results on FDDB against the recent state-of-the-art face detection methods.
%----------
\end{abstract}
}

\maketitle

% is not selected <OR> if conference mode is selected - because compsoc
% conference papers position the abstract like regular (non-compsoc)
\IEEEdisplaynotcompsoctitleabstractindextext

% \fi
\IEEEpeerreviewmaketitle

%-------------------------------------------------------------------------
%%%%%%%%% BODY TEXT
%\vspace{-10mm}
\section{Introduction}

Detection and analysis on human subjects using facial feature based biometrics for access control, surveillance systems and other security applications have gained popularity over the past few years. Several such biometrics systems are deployed in security checkpoints across the globe with more being deployed every day.
Particularly, face recognition has been one of the most popular biometrics modalities attractive to security departments. Indeed, the uniqueness of facial features across individuals can be captured much more easily than other biometrics. In order to take into account a face recognition algorithm, however, face detection usually needs to be done first.
\begin{figure}[!t]
	\centering \includegraphics[width=\columnwidth]{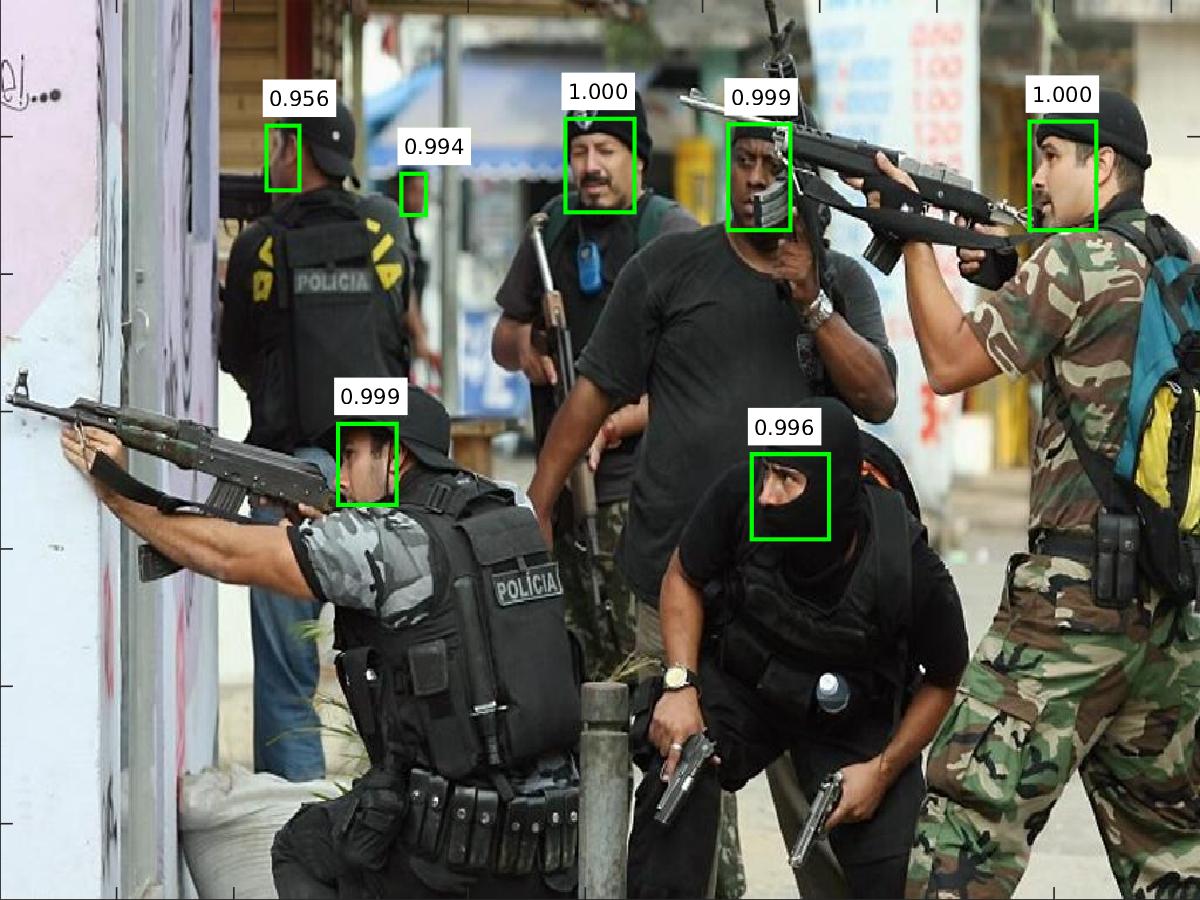}
		\caption{An example of face detection results using our proposed CMS-RCNN method. The proposed method can robustly detect faces across occlusion, facial expression, pose, illumination, scale and low resolution conditions from WIDER FACE Dataset \cite{yang2016wider}. }
	\label{fig:face_det_1}
\end{figure}

The problem of face detection has been intensely studied for decades with the aim of ensuring the generalization of robust algorithms to unseen face images \cite{viola2001rapid,zhang2010survey,zhu2012face,li2013learning,li2013pep-adapt,markuvs2013pico,li2014exemplar,mathias2014face,chen2014jointcascade,yang2014acf-multiscale,ghiasi2015multireshpm,liao2014npdface}. Although the detection accuracy in recent face detection algorithms \cite{li2015cascadecnn,farfade2015ddfd,yang2015faceness,ranjan2015dp2mfd,yang2015ccf,ranjan2016hyperface} has been highly improved due to the advancement of deep Convolutional Neural Networks (CNN), they are still far from achieving the same detection capabilities as a human due to a number of challenges in practice. For example, off-angle faces, large occlusions, low-resolutions and strong lighting conditions, as shown in Figure \ref{fig:face_det_1}, are always the important factors that need to be considered.

This paper presents an advanced CNN based approach named Contextual Multi-Scale Region-based CNN (CMS-RCNN) to handle the problem of face detection in digital face images collected under numerous challenging conditions, e.g. heavy facial occlusion, illumination, extreme off-angle, low-resolution, scale difference, etc. Our designed region-based CNN architecture allows the network to simultaneously look at multi-scale features, as well as to explicitly look outside facial regions as the potential body regions. In other words, this process tries to mimic the way of face detection by human in a sense that when humans are not sure about a face, seeing the body will increase our confidence. Additionally this architecture also helps to synchronize both the global semantic features in high level layers and the localization features in low level layers for facial representation.
Therefore, it is able to robustly deal with the challenges in the problem of unconstrained face detection.

\begin{figure}
\centering
\includegraphics[width=\columnwidth]{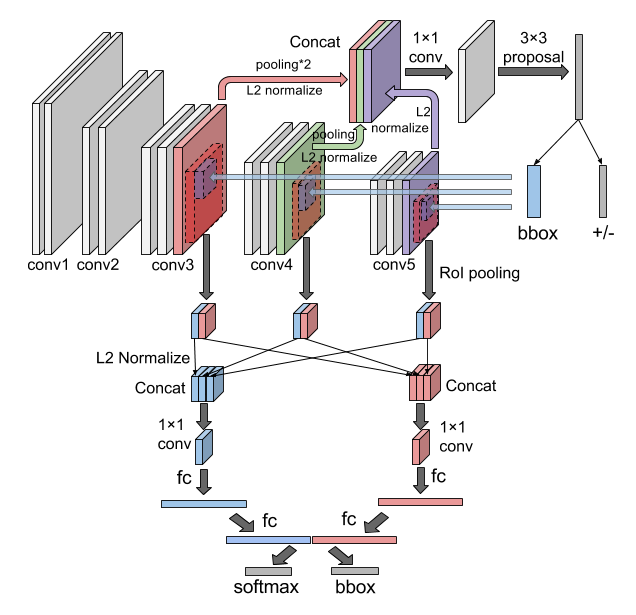}
\caption{Our proposed Contextual Multi-Scale Region-based CNN model. It is based on the VGG-16 model \cite{simonyan2014very}, with 5 sets of convolution layers in the middle. The upper part is the Multi-Scale Region Proposal Network (MS-RPN) and the lower part is the Contextual Multi-Scale Convolution Neural Network (CMS-CNN). In the CMS-CNN, the face features labeled as blue blocks and the body context features labeled as red blocks are processed in parallel and combined in the end for final outputs, i.e. confidence score and bounding box.}
\label{fig:msfrcnn_fw}
\end{figure}

Our CMS-RCNN method introduces the Multi-Scale Region Proposal Network (MS-RPN) to generate a set of region candidates and the Contextual Multi-Scale Convolution Neural Network (CMS-CNN) to do inference on the region candidates of facial regions. A confidence score and bounding box regression are computed for every candidate.
In the end, the face detection system is able to decide the quality of the detection results by thresholding these generated confidence scores in given face images.
The architecture of our proposed CMS-RCNN network for unconstrained face detection is illustrated in Figure~\ref{fig:msfrcnn_fw}.

Our approach is evaluated on two challenging face detection databases and compared against numerous recent face detection methods. Firstly, the proposed CMS-RCNN method is compared against four strong baselines \cite{yang2014acf-multiscale,yang2015faceness,yang2016wider} on the WIDER FACE Dataset \cite{yang2016wider}, a large scale face detection benchmark database. This experiment shows its capability to detect face images in the wild, e.g. under occlusions, illumination, facial poses, low-resolution conditions, etc. Our method outperforms the baselines by a huge margin in all easy, medium, and hard partitions.
It is also benchmarked on the Face Detection Data Set and Benchmark (FDDB) \cite{fddbTech}, a dataset of face regions designed for studying the problem of unconstrained face detection.
The experimental results show that the proposed CMS-RCNN approach consistently achieves highly competitive results against the other state-of-the-art face detection methods. 
% Finally, we present the limitations of the proposed MS-FRCNN method in the problem of face detection.

The rest of this paper is organized as follows. In section \ref{sec:related}, we summarize prior work in face detection. Section \ref{sec:background} reviews a general deep learning framework, the background as well as the limitations of the Faster R-CNN in the problem of face detection. In Section \ref{sec:Ourapproach}, we introduce our proposed CMS-RCNN approach for the problem of unconstrained face detection. Section \ref{sec:ExptsRes} presents the experimental face detection results and comparisons obtained using our proposed approach on two challenging face detection databases, i.e. the WIDER FACE Dataset and the FDDB database. Finally, our conclusions in this work are presented in Section \ref{sec:Concl}.

%========================================================================

\section{Related Work}
\label{sec:related}

% Write something to introduce the related work.

% \subsection{Face Detection}
% \label{subsec:facedec}

Face detection has been a well studied area of computer vision. One of the first well performing approaches to the problem was the Viola-Jones face detector \cite{viola2001rapid}. It was capable of performing real time face detection using a cascade of boosted simple Haar classifiers. The concepts of boosting and using simple features has been the basis for many different approaches \cite{zhang2010survey} since the Viola-Jones face detector. These early detectors tended to work well on frontal face images but not very well on faces in different poses. As time has passed, many of these methods have been able to deal with off-angle face detection by utilizing multiple models for the various poses of the face. This increases the model size but does afford more practical uses of the methods. Some approaches have moved away from the idea of simple features but continued to use the boosted learning framework. Li and Zhang \cite{li2013learning} used SURF cascades for general object detection but also showed good results on face detection.

More recent work on face detection has tended to focus on using different models such as a Deformable Parts Model (DPM) \cite{zhu2012face, Felzenszwalb}. Zhu and Ramanan's work was an interesting approach to the problem of face detection where they combined the problems of face detection, pose estimation, and facial landmarking into one framework. By utilizing all three aspects in one framework, they were able to outperform the state-of-the-art at the time on real world images. Yu et al. \cite{yu2013pose} extended this work by incorporating group sparsity in learning which landmarks are the most salient for face detection as well as incorporating 3D models of the landmarks in order to deal with pose. Chen et al. \cite{chen2014jointcascade} have combined ideas from both of these approaches by utilizing a cascade detection framework while simultaneously localizing features on the face for alignment of the detectors. Similarly, Ghiasi and Fowlkes \cite{ghiasi2015multireshpm} have been able to use heirarchical DPMs not only to achieve good face detection in the presence of occlusion but also landmark localization. However, Mathias et al. \cite{mathias2014face} were able to show that both DPM models and rigid template detectors similar to the Viola-Jones detector have a lot of potential that has not been adequately explored. By retraining these models with appropriately controlled training data, they were able to create face detectors that perform similarly to other, more complex state-of-the-art face detectors.

All of these approaches to face detection were based on selecting a feature extractor beforehand. However, there has been work done in using a ConvNet to learn which features are used to detect faces. Neural Networks have been around for a long time but have been experiencing a resurgence in popularity due to hardware improvements and new techniques resulting in the capability to train these networks on large amounts of training data. Li et al. \cite{li2015cascadecnn} utilized a cascade of CNNs to perform face detection. The cascading networks allowed them to process different scales of faces at different levels of the cascade while also allowing for false positives from previous networks to be removed at later layers in a similar approach to other cascade detectors. Yang et al. \cite{yang2015faceness} approached the problem from a different perspective more similar to a DPM approach. In their method, the face is broken into several facial parts such as hair, eyes, nose, mouth, and beard. By training a detector on each part and combining the score maps intelligently, they were able to achieve accurate face detection even under occlusions. Both of these methods require training several networks in order to achieve their high accuracy. Our method, on the other hand, can be trained as a single network, end-to-end, allowing for less annotation of training data needed while maintaining highly accurate face detection.

The ideas of using contextual information in object detection have been studied in several recent work with very high detection accuracy. 
Divvala et al. \cite{divvala2009empirical} reviewed the the role of context in a contemporary, challenging object detection in their empirical evaluation
analysis. In their conclusions, the context information not only reduces the overall detection errors, but also the remaining errors made by the detector are more reasonable. Bell et al. \cite{bell2015inside} introduced an advanced object detector method named Inside-Outside Network (ION) to exploit information both inside and outside the region of interest. In their approach, the contextual information outside the region of interest is incorporated using spatial recurrent neural networks. Inside the network, skip pooling is used to extract information at multiple scales and levels of abstraction. Recently, Zagoruyko et al. \cite{zagoruyko2016multipath} have presented 
the MultiPath network with three modifications to the standard Fast R-CNN object detector, i.e. skip connections that give the detector access to features at multiple network layers, a foveal structure to exploit object context at multiple object resolutions, and an integral loss function and corresponding network adjustment that improve localization. The information in their proposed network can flow along multiple paths. Their MultiPath network is combined with DeepMask object proposals to solve the object detection problem.

Unlike all the previous approaches that select a feature extractor beforehand and incorporate a linear classifier with the depth descriptor beside RGB channels, our method solves the problem under a  deep learning framework where the global and the local context features, i.e. multi scaling, are synchronized to Faster Region-based Convolutional Neural Networks in order to robustly achieve semantic detection.

%========================================================================

\section{Background}
\label{sec:background}

The recent studies in deep ConvNets have achieved significant results in object detection, classification and modeling \cite{krizhevsky2012imagenet}. In this section, we review various well-known Deep ConvNets. Then, we show the current limitations of the Faster R-CNN, one of the state-of-the-art deep ConvNet methods in object detection, in the defined context of the face detection.

% \subsection{Deep Learning Framework}
% \label{subsec:DLframework}

% Convolutional Neural Networks are biologically inspired variants of multilayer perceptrons. The ConvNet method and its extensions, e.g. LeNet-5, HMAX, etc., imitate the characteristics of animal visual cortex systems that contain a complex arrangement of cells sensitive to receptive fields. In their models, the designed filters are considered as human visual cells in order to explore spatially local correlations in natural images.
% It efficiently presents the sparse connectivity and the shared weights since these kernel filters are replicated over the entire image with the same parameters in each layer. In addition, the pooling step, a form of down-sampling, plays a key role in ConvNet. Max-pooling is a popular pooling method for object detection and classification since max-pooling reduces computation for upper layers by eliminating non-maximal values and provides a small amount of translation invariance in each level.

% Although ConvNets can explore deep features, they are very computationally expensive. The algorithm becomes more practical when implemented in a Graphics Processing Unit (GPU). The Caffe framework \cite{jia2014caffe} is one of the fastest deep learning implementations using CUDA C++ for GPU computation. It also supports interfaces to Python/Numpy and MATLAB. It can be used as an off-the-shelf deployment of the state-of-the-art models.
% This framework is employed in our implementation.

\subsection{Region-based Convolution Neural Networks}
One of the most important approaches for the object detection task is the family of Region-based Convolution Neural Networks (R-CNN). 

R-CNN \cite{girshick2016region}, the first generation of this family, applies the high-capacity deep ConvNet to classify given bottom-up region proposals. Due to the lack of labeled training data, it adopts a strategy of supervised pre-training for an auxiliary task followed by domain-specific fine-tuning. Then the ConvNet is used as a feature extractor and the system is further trained for object detection with Support Vector Machines (SVM). Finally, it performs bounding-box regression. The method achieves high accuracy but is very time-consuming. The system takes a long time to generate region proposals, extract features from each image, and store these features in a hard disk, which also takes up a large amount of space. At testing time, the detection process takes 47s per image using VGG-16 network \cite{simonyan2014very} implemented in GPU due to the slowness of feature extraction. In other words, R-CNN is slow because it processes each object proposal independently without sharing computation. 

Fast R-CNN \cite{girshick2015fast} solves this problem by sharing the features between proposals. The network is designed to only compute a feature map once per image in a fully convolutional style, and to use ROI-pooling to dynamically sample features from the feature map for each object proposal. The network also adopts a multi-task loss, i.e. classification loss and bounding-box regression loss. Based on the two improvements, the framework is trained end-to-end. The processing time for each image significantly reduced to 0.3s. Fast R-CNN accelerates the detection network using the ROI-pooling layer. However the region proposal step is designed out of the network hence still remains a bottleneck, which results in sub-optimal solution and dependence on the external region proposal methods. 

Faster R-CNN \cite{ren2015faster} addresses the problem with fast R-CNN by introducing the Region Proposal Network (RPN). An RPN is implemented in a fully convolutional style to predict the object bounding boxes and the objectness scores. In addition, the anchors are defined with different scales and ratios to achieve the translation invariance. The RPN shares the full-image convolution features with the detection network. Therefore the whole system is able to complete both proposal generation and detection computation within 0.2s using very deep VGG-16 model \cite{simonyan2014very}. With a smaller ZF model \cite{zeiler2014visualizing}, it can reach the level of real-time processing.

\subsection{Limitations of Faster R-CNN}

The Region-based CNN family, e.g. Faster R-CNN and its variants \cite{girshick2015fast}, achieves the state-of-the-art performance results in object detection on the PASCAL VOC dataset. These methods can detect objects such as vehicles, animals, people, chairs, and etc. with very high accuracy.
In general, the defined objects often occupy the majority of a given image. However, when these methods are tested on the challenging Microsoft COCO dataset \cite{lin2014microsoft}, the performance drops a lot, since images contain more small, occluded and incomplete objects. Similar situations happen in the problem of face detection. We focus on detecting only facial regions that are sometimes small, heavily occluded and of low resolution (as shown in Figure \ref{fig:face_det_1}).

The detection network in designed Faster R-CNN is unable to robustly detect such tiny faces. The intuition point is that the Regions of Interest pooling layer, i.e. ROI-pooling layer, builds features only from the last single high level feature map.
For example, the global stride of the 'conv5' layer in the VGG-16 model is 16. Therefore, given a facial region with the sizes less than $16\times16$ pixels in an image, the projected ROI-pooling region for that location will be less than 1 pixel in the 'conv5' layer, even if the proposed region is correct. Thus, the detector will have much difficulty to predict the object class and the bounding box location based on information from only one pixel.

% - Do we need to add some background about MS-FRCNN foreshadowing the next subsection? We lack for continuity here. 

% \subsection{When MS-FRCNN Fails in Face Detection}
% \label{subsec:exp_fails}

% In Wider Face database, there are many tiny labeled facial regions that need to be learned. The proposed method is trained on too many faces in those low quality conditions. Indeed, the human facial features in those facial regions are very limited.
% Therefore, the algorithm over-fits deep facial features in some cases. Given a new testing image, the trained system may mislabel some small regions with complicated patterns as human faces as shown in Figure \ref{fig:failure}. This is the point we will explore for a better solution in future.

\subsection{Other Face Detection Method Limitations}

Other challenges in object detection in the wild include occlusion and low-resolution. For face detection, it is very common for people to wear stuffs like sunglasses, scarf and hats, which occlude the face. In such cases, the methods that only extract features from faces do not work well. For example, Faceness \cite{yang2015faceness} consider finding faces through scoring facial parts responses by their spatial structure and arrangement, which works well on clear faces. But when facial parts are missing due to occlusion or when face itself is too small, facial parts become more hard to detect. Therefore, the body context information plays its role. As an example of context-dependent objects, faces often come together with human body. Even though the faces are occluded, we can still locate it only by seeing the whole human body. Similar advantages for faces at low-resolution, i.e. tiny faces. The deep features can not tell much about tiny faces since their receptive field is too small to be informative. Introducing context information can extend the area to extract features and make them meaningful. On the other hand, the context information also helped with reducing false detection as discussed previously, since context information tells the difference between real faces with bodies and face-like patterns without bodies. 

%Mention about:

%- Context information, i.e. when face or hand unclear, very low-resolution, these methods cannot detect. Semantically, human usually borrow the information beyond facial information, i.e. body structures, in order to predict the facial regions.

%- This context also help to reduce the false detection when having any object similar to faces.

%- Deep feature, small receptive field is not good to detect hands or faces with small sizes.

%========================================================================
% \begin{figure}[!t]
% \centering
% \includegraphics[width=0.9\columnwidth]{Figs/MS_vs_FRCNN.png}
% \caption{The face detection comparison between our proposed MS-FRCNN and the Faster R-CNN on the Wider Face validation set \cite{yang2015widerface}.}
% \label{fig:ms_vs_frcnn}
% \end{figure}

\section{Contextual Multi-Scale R-CNN}
\label{sec:Ourapproach}
Our goal is to detect human faces captured under various challenging conditions such as strong illumination, heavily occlusion, extreme off-angles, and low resolution. Under these conditions, the current CNN-based detection systems suffer from two major problems, i.e. 1) tiny faces are hard to identify; 2) only face region is taken into consideration for classification. In this section, we show why these problems hinder the ability of a face detection system. Then, our proposed network is presented to address these problems by using the Multi-Scale Region Proposal Network (MS-RPN) and the Contextual Multi-Scale Convolution Neural Network (CMS-CNN), as illustrated in Figure~\ref{fig:msfrcnn_fw}. Similar to Faster R-CNN, the MS-RPN outputs several region candidates and the CMS-CNN computes the confidence score and bounding box for each candidate.

% Current state-of-the-art detection methods have limitations to detect under extreme conditions, mainly due to two drawbacks in the network design, 
% This section presents our proposed Contextual Multi-Scale R-CNN approach to robustly detect facial regions. Our approach utilizes the deep features encoded in both the global and the local representation for facial regions. Since the values of the filter responses range in different scales in each layer, i.e. the deeper a layer is, the smaller values of the filter responses are, there is a need for a further calibration process to synchronize the values received from multiple filter responses. The average feature for layers in Faster-RCNN are employed to augment features at each location.

\subsection{Identifying Tiny Faces}
Why tiny faces are hard to be robustly detected by the previous region-based CNNs? The reason is that in these networks both the proposed region and the classification score are produced from one single high-level convolution feature map. This representation doesn't have enough information for the multiple tasks, i.e. region proposal and RoI detection. 
For example, Faster R-CNN generates region candidates and does RoI-pooling from the 'conv5' layer of the VGG-16 model, which has a overall stride of 16. One issue is that the reception field in this layer is quite large. When the face size is less than 16-by-16 pixels, the corresponding output in 'conv5' layer is less than 1 pixel, which is insufficient to encode informative features. The other issue is that as the convolution layers go deeper, each pixel in the feature map gather more and more information outside the original input region so that it contains lower proportion of information for the region of interest. These two issues together make the last convolution layer less representative for tiny faces.

\subsubsection{Multiple Scale Faster-RCNN}
Our solution for this problem is a combination of both global and local features, i.e. multiple scales. In this architecture, the feature maps are incorporated from lower level convolution layers with the last convolution layer for both MS-RPN and CMS-CNN. Features from lower convolution layer help get more information for the tiny faces, because stride in lower convolution layer will not be too small. Another benefit is that both low-level feature with localization capability and high-level feature with semantic information are fused together \cite{hariharan2015hypercolumns}, since face detection needs to localize the face as well as to identify the face.
In the MS-RPN, the whole lower level feature maps are down-sampled to the size of high level feature map and then concatenated with it to form a unified feature map. Then we reduce the dimension of the unified feature map and use it to generate region candidates. 
In the CMS-CNN, the region proposal is projected into feature maps from multiple convolution layers. And RoI-pooling is performed in each layer, resulting in a fixed-size feature tensor. All feature tensors are normalized, concatenated and dimension-reduced to a single feature blob, which is forwarded to two fully connected layers to compute a representation of the region candidate.

\subsubsection{L2 Normalization}
In both MS-RPN and CMS-CNN, concatenation of feature maps is done with L2 normalization layer \cite{liu2015parsenet}, shown in Fig. \ref{fig:msfrcnn_fw}, since the feature maps from different layer have generally different properties in terms of numbers of channels, scale of value and norm of feature map pixels. Generally, comparing with values in shallower layers, the values in deeper layers are usually too small, which leads to the dominance of shallower layers. In practice, it is impossible for the system to readjust and tune value from each layer for best performance. Therefore, L2 normalization layers before concatenation are crucial for the robustness of the system because it keeps the value from each layer in roughly the same scale. 

The normalization is performed within each pixel, and all feature map is treated independently:

\[
\begin{gathered}
\hat{\mathbf{x}} = \frac{\mathbf{x}}{\left \| \mathbf{x} \right \|_2} \\
\left \| \mathbf{x} \right \|_2 = ( \sum_{i=1}^{d}{\left | x_i \right |} ) ^ {\frac{1}{2}}
\end{gathered}
\]
where the $\mathbf{x}$ and $\hat{\mathbf{x}}$ stand for the original pixel vector and the normalized pixel vector respectively. $d$ stands for the number of channels in each feature map tensor. 

During training, scaling factors $\gamma_i$ will be updated to readjust the scale of the normalized features. For each channel $i$, the scaling factor follows: 
\[
\begin{gathered}
y_i=\gamma_i \hat{{x}}_i
\end{gathered}
\]
where $y_i$ stand for the re-scaled feature value. 

Following the back-propagation and chain rule, the update for scaling factor $\gamma$ is:
\[
\begin{split}
& \frac{\partial l}{\partial \hat{\mathbf{x}}}=\frac{\partial l}{\partial \mathbf{y}}\cdot \gamma \\
& \frac{\partial l}{\partial \mathbf{x}}=\frac{\partial l}{\partial \hat{\mathbf{x}}}\left ( \frac{\mathbf{I}}{\left \| \mathbf{x} \right \|_2}-\frac{\mathbf{x}\mathbf{x}^T}{\left \| \mathbf{x} \right \|_2^3} \right ) \\
& \frac{\partial l}{\partial \gamma_i}=\sum_{y_i}{\frac{\partial l}{\partial y_i}\hat{x}_i}
\end{split}
\]
where $\mathbf{y}=\left[ y_1,y_2,...,y_d\right]^T$. 

\subsubsection{New Layer in Deep Learning Caffe Framework}
\label{subsec:ms_caffe}

%The implementation of L2 normalization is similar to the work of ParseNet \cite{liu2015parsenet}. 
The system integrate information from lower layer feature maps, i.e. third and fourth convolution layers, to extract determinant features for tiny faces. For both parts of our system, i.e. MS-RPN and CMS-CNN, the L2 normalization layers are inserted before concatenation of feature maps from the three layers. The features were re-scaled to proper values and concatenated to a single feature map. We set the initial scaling factor in a special way, following two rules. First, the average scale for each feature map is roughly identical; second, after the following $1\times 1$ convolution, the resulting tensor should have the same average scale as the conv5 layer in the work of Faster R-CNN. As implied, after the following $1\times 1$ convolution, the tensor should be the same as the original architecture in Faster R-CNN, in terms of its size, scale of values and function for the downstream process.  

\subsection{Integrating Body Context}
\label{subsec:context}
\begin{figure}
\centering
\includegraphics[height=3cm]{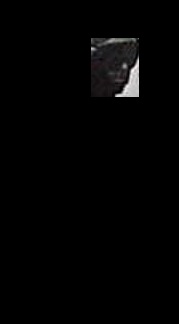}
\includegraphics[height=3cm]{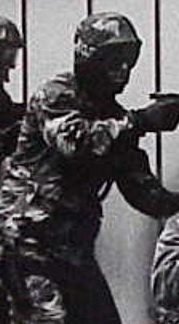}
\includegraphics[height=3cm]{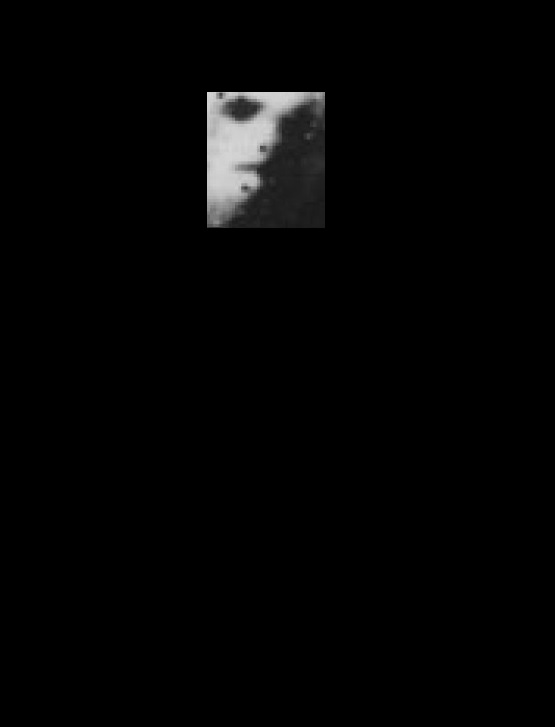}
\includegraphics[height=3cm]{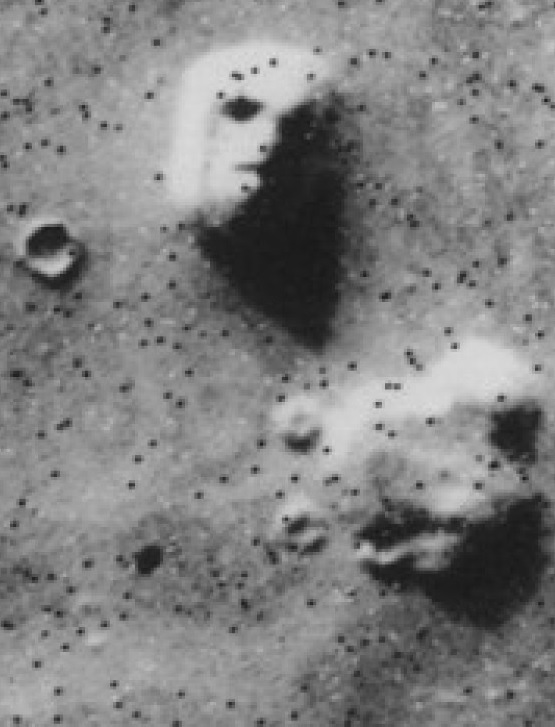}
\caption{Examples of body context helping face identification. The first two figures show that existence of a body can increase the confidence of finding a face. The last two figures show that what looks like a face turns out to be a mountain on the planet surface when we see more context information.}
\label{fig:body_context}
\end{figure}

When humans are searching for faces, they try to look for not only the facial patterns, e.g. eyes, nose, mouth, but also the human bodies. Sometimes a human body makes us more convinced about the existence of a face. In addition, sometimes human body helps to reject false positives. If we only look at face regions, we may make mistakes identifying them. For example, Figure~\ref{fig:body_context} shows two cases where body region plays a significant role for correct detection. This intuition is not only true for human but also valid in computer vision. Previous research has shown that contextual reasoning is a critical piece of the object recognition puzzle, and that context not only reduces the overall detection errors, but, more importantly, the remaining errors made by the detector are more reasonable \cite{divvala2009empirical}. Based on this intuition, our network is designed to make explicit reference to the human body context information in the RoI detection.

% Faces can be extremely hard to detect in the cases of low resolution, off-angle, blurred or occluded. Few face detector can find the face by only rely on extracting features from the face region. Human, on the other hand, can easily tell a face in most of the extreme cases discussed above since we also incorporate knowledge from the structure of human body. That is, if there is a human body, there must be a face. Also  Therefore, inspired by this intuition, we integrate context information in our proposed system. 

%The way to introduce the context information to the system in our implementation is by adding an extra ROI-pooling layer 

In our proposed network, the contextual body reasoning is implemented by explicitly grouping body information from convolution feature maps shown as the red blocks in Figure~\ref{fig:msfrcnn_fw}. Specifically, additional RoI-pooling operations are performed for each region proposal in convolution feature maps to represent the body context features. Then same as the face feature tensors, these body feature tensors are normalized, concatenated and dimension-reduced to a single feature blob. After two fully connected layers the final body representation is concatenated with the face representation. They together contribute to the computation of confidence score and bounding box regression.

\begin{figure}
    \centering
    \includegraphics[width=\columnwidth]{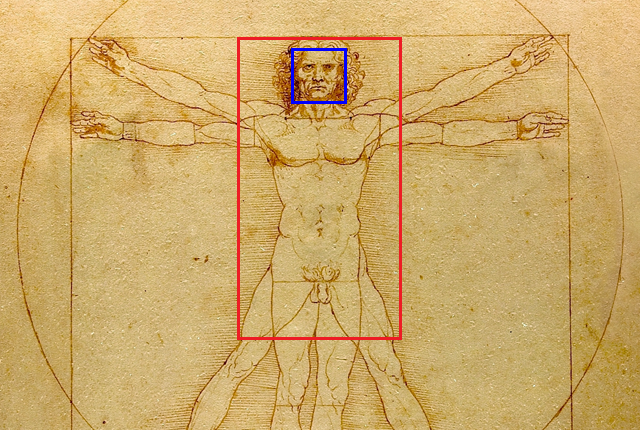}
    \caption{The Vitruvian Man: spatial relation between the face (blue box) and the body (red box).}
    \label{fig:man}
\end{figure}

With projected region proposal as the face region, the additional RoI-pooling region represents the body region and satisfies a pre-defined spatial relation with the face region. In order to model this spatial relation, we make a simple hypothesis that if there is a face, there must exist a body, and the spatial relation between each face and body is fixed. This assumption may not be true all the time but should cover most of the scenarios since most people we see in the real world are either standing or sitting. Therefore, the spatial relation is roughly fixed between the face and the vertical body. 
% The relation between  In the real world, we usually see people standing or sitting. In either case the body is vertical. So this spatial relation between face and body has covered most of the scenarios. 
Mathematically, this spatial relation can be represented by four parameters presented in Equation~\ref{eq:spatial_rel}. 
\begin{equation}
\label{eq:spatial_rel}
    \begin{split}
        t_x &= (x_b - x_f)/w_f \\
        t_y &= (y_b - y_f)/h_f \\
        t_w &= \log(w_b/w_f) \\
        t_h &= \log(h_b/h_f) \\
    \end{split}
\end{equation}
where $x_{(*)}$, $y_{(*)}$, $w_{(*)}$, and $h_{(*)}$  denote the two coordinates of the box center, width, and height respectively. And $b$ and $f$ stand for body and face respectively. $t_x$, $t_y$, $t_w$, and $t_h$ are the parameters. Through out this paper, we fix the for parameters such that the two projected RoI regions of face and body satisfies a certain spatial ratio illustrated in the famous drawing in Figure~\ref{fig:man}.

\subsection{Information Fusion}
\label{subsec:info_fuse}
% talk about l2 normalization and late fusion v.s. early fusion
It's worth noticing that in our deep network architecture we have multiple face feature maps and body context feature maps for each proposed region. A critical issue is how we effectively fuse these information, i.e. what computation to apply and in which stage.

% Now that we have , we need to fuse them to get a uniform representation. We first normalize each feature map along the channel dimension to make them have the same scale. This is crucial because the scales of feature from different level of layers can be very different\cite{liu2015parsenet}. Naively concatenating them will cause the domination of the feature map with the highest scale. After normalization, all feature maps are concatenated and go through dimension reduction. The resulting feature map is feeded into a fully convolution layer followed by the traditional multi-task learning of classification and bounding box regression.
% \subsection{Multiple Scale Normalization}
% \label{subsec:ms_l2norm}

% Khoa/Yutong rewrite ...

In our network, features extracted from different convolution layers need to be fused together to get a uniform representation. They cannot be naively concatenated due to the overall differences of the numbers of channels, scales of values and norms of feature map pixels among these layers. The detailed research shows that the deeper layers often contain smaller values than the shallower layers. Therefore, the larger values will dominate the smaller ones, making the system rely too much on shallower features rather than a combination of multiple scale features causing the system to no longer be robust. We adopt the normalization layer from \cite{liu2015parsenet} to address this problem. The system takes the multiple scale features and apply L2 normalization along the channel axis of each feature map. Then, since the channel size is different among layers, the normalized feature map from each layer needed to be re-weighted, so that their values are at the same scale. After that, the feature maps are concatenated to one single feature map tensor. This modification helps to stabilize the system and increase the accuracy. Finally, the channel size of the concatenated feature map is shrunk to fit right in the original architecture for the downstream fully-connected layers.

Another crucial question is whether to fuse the face information and the body information at a early stage or at the very end of the network. Here we choose the late fusion strategy in which face features and body context features are extracted in two parallel pipelines. At the very end of the network two representations for face and body context are concatenated together to form a long feature vector. Then this feature vector is forwarded to compute confidence score and bounding box regression. The other strategy is the early fusion, in which face feature maps and body context feature maps get concatenated right after RoI pooling and normalization. These two strategies both combine the information from face and body context, but we prefer the late fusion. The reason is that we want the network to make decisions in a more semantic space. We care more about the existence of the face and the body. The localization information is already encoded in the predefined spatial relation mentioned in Section~\ref{subsec:context}. Moreover empirical experiments also show that late fusion strategy works better.

% \subsection{Training}
% \label{subsec:training}

% Describe how to train the system,
% HOw much time cost for training
% Compare between

\subsection{Implementation Details}
\label{subsec:details}
Our CMS-RCNN is implemented in the Caffe deep learning framework \cite{jia2014caffe}. The first 5 sets of convolution layers have the same architecture as the deep VGG-16 model, and during training their parameters are initialized from the pre-trained VGG-16. For simplicity we refer to the last convolution layers in set 3, 4 and 5 as 'conv3', 'conv4', and 'conv5' respectively. All the following layers are connected exclusively to these three layers.
In the MS-RPN, we want 'conv3', 'conv4', and 'conv5' to be synchronized to the same size so that concatenation can be applied. So 'conv3' is followed by pooling layer to perform down-sampling. Then 'conv3', 'conv4', and 'conv5' are normalized along the channel axis to a learnable re-weighting scale and concatenated together. To ensure training convergence, the initial re-weighting scale needs to be carefully set. Here we set the initial scale of 'conv3', 'conv4', and 'conv5' to be 66.84, 94.52, and 94.52 respectively. 
In the CMS-CNN, the RoI pooling layer already ensure that the pooled feature maps have the same size. Again we normalize the pooled features to make sure the downstream values are at reasonable scales when training is initialized. Specifically, features pooled from 'conv3', 'conv4', and 'conv5' are initialized with scale to be 57.75, 81.67, and 81.67 respectively, for both face and body pipelines.
The MS-RPN and the CMS-CNN share the same parameters for all convolution layers so that computation can be done once, resulting in higher efficiency. Additionally, in order to shrink the channel size of the concatenated feature map, a $1\times 1$ convolution layer is then employed. Therefore the channel size of final feature map is at the same size as the original fifth convolution layer in Faster R-CNN.

\begin{figure*}[ht]
\centering
\subfigure[]
{
    \includegraphics[height=5.0cm, width=5.4cm]{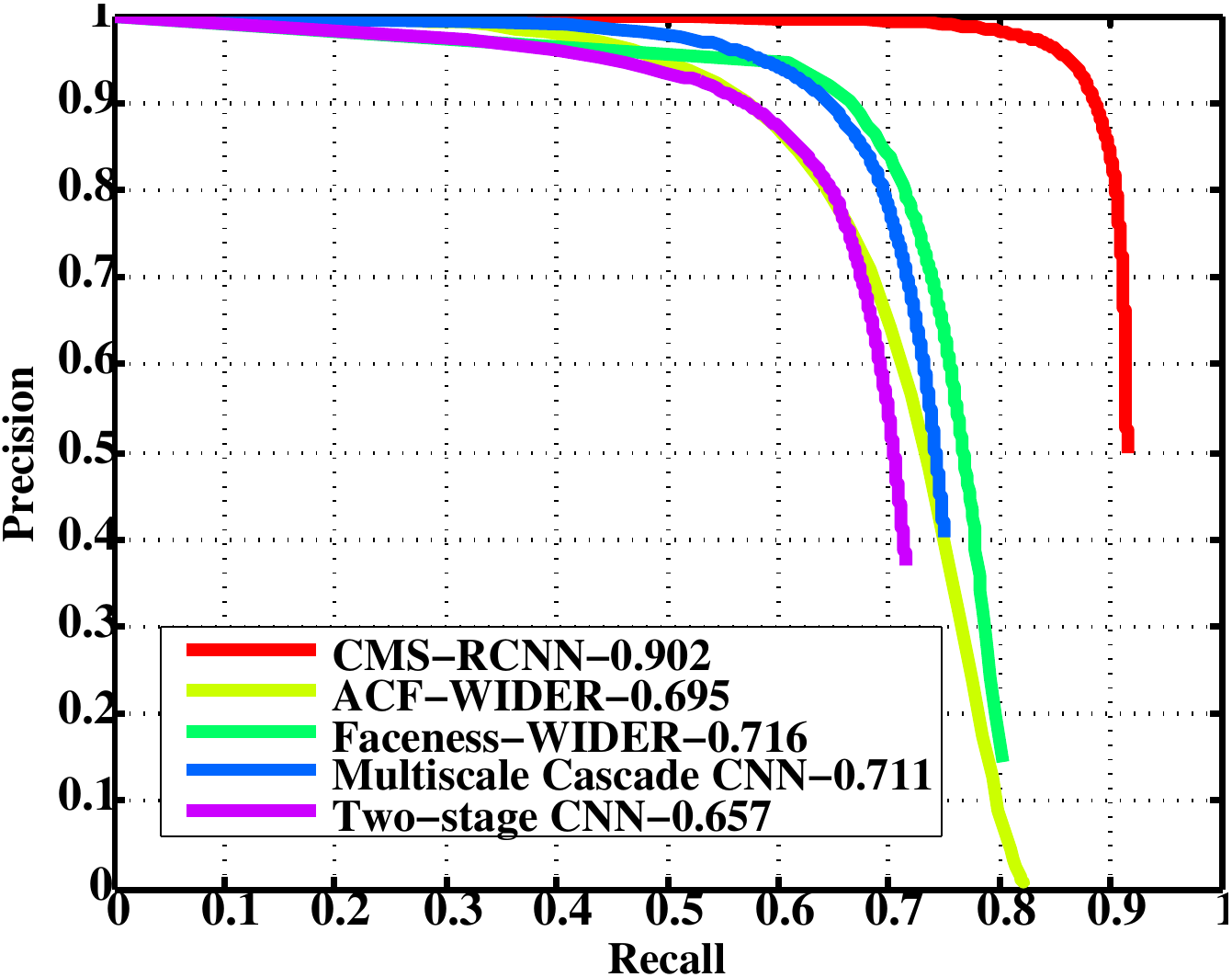}
}
\hspace{-5mm}
\subfigure[]
{
    \includegraphics[height=5.0cm, width=5.4cm]{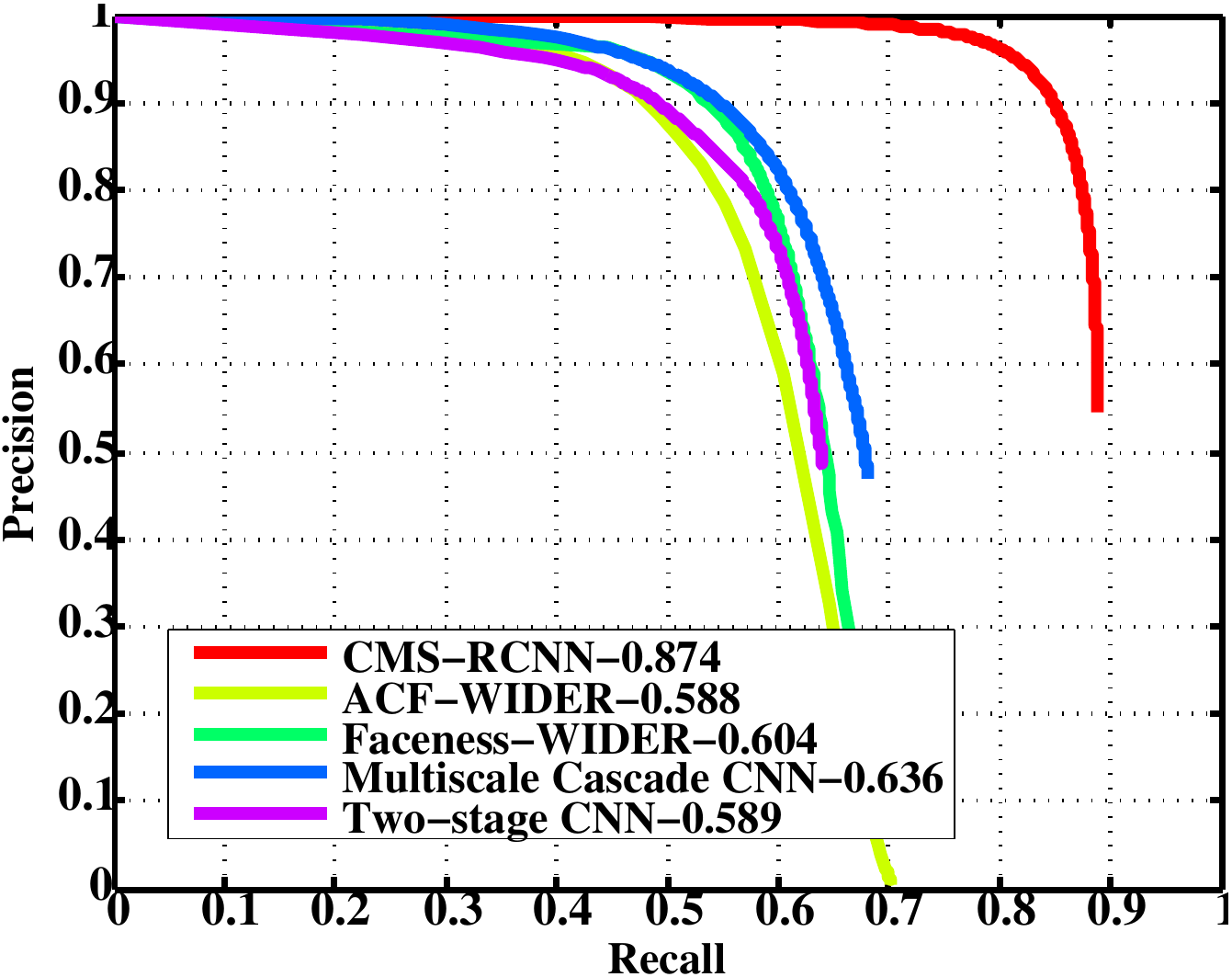}
}
\hspace{-5mm}
\subfigure[]
{
    \includegraphics[height=5.0cm, width=5.4cm]{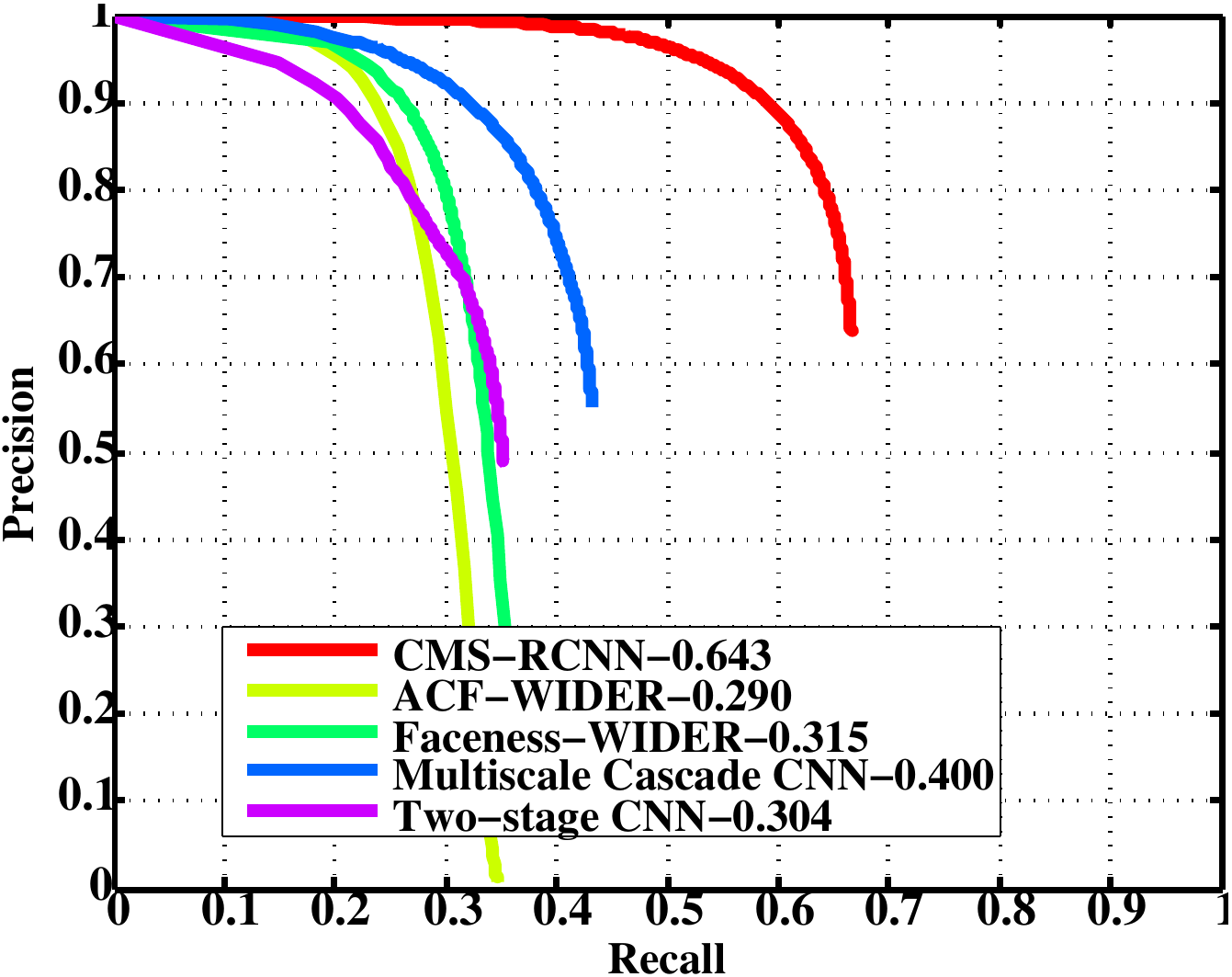}
}
\hspace{-5mm}
\caption{Precision-Recall curves obtained by our proposed CMS-RCNN (red) and the other baselines, i.e.
Two-stage CNN \cite{yang2016wider}, Multi-scale Cascade CNN \cite{yang2016wider}, Faceness \cite{yang2015faceness}, and Aggregate Channel Features (ACF) \cite{yang2014acf-multiscale}. All methods trained and tested on the same training and testing set of the WIDER FACE dataset. (a): Easy level, (b): Medium level and (c): Hard level. Our method achieves the state-of-the-art results with the highest AP values of 0.902 (Easy), 0.874 (Medium) and 0.643 (Hard) among the methods on this database. It also outperforms the second best baseline by 26.0\% (Easy), 37.4\% (Medium) and 60.8\% (Hard).}
\label{fig:roc_wider}
\end{figure*}

\section{Experiments}
\label{sec:ExptsRes}

This section presents the face detection bechmarking using our proposed CMS-RCNN approach on the WIDER FACE dataset \cite{yang2016wider} and the Face Detection Data Set and Benchmark (FDDB) \cite{fddbTech} database. The WIDER FACE dataset is experimented with high degree of variability. Using this database, our proposed approach robustly outperforms strong baseline methods, including Two-stage CNN \cite{yang2016wider}, Multi-scale Cascade CNN \cite{yang2016wider}, Faceness \cite{yang2015faceness} and Aggregate Channel Features (ACF) \cite{yang2014acf-multiscale}, by a large margin. 
We also show that our model trained on WIDER FACE dataset generalizes well enough to the FDDB database. The trained model consistently achieves competitive results against the recent state-of-the-art face detection methods on this database, including HyperFace \cite{ranjan2016hyperface}, DP2MFD \cite{ranjan2015dp2mfd}, CCF \cite{yang2015ccf}, Faceness \cite{yang2015faceness}, NPDFace \cite{liao2014npdface}, MultiresHPM \cite{ghiasi2015multireshpm}, DDFD \cite{farfade2015ddfd}, CascadeCNN \cite{li2015cascadecnn}, ACF-multiscale \cite{yang2014acf-multiscale}, Pico \cite{markuvs2013pico}, HeadHunter \cite{mathias2014face}, Joint Cascade \cite{chen2014jointcascade}, Boosted Exemplar \cite{li2014exemplar}, and PEP-Adapt \cite{li2013pep-adapt}. 
% In Section \ref{subsec:exp_training}, we present the training steps on the Wider Face database.
% In Section \ref{subsec:ms_vs_frcnn}, the face detection results using MS-FRCNN and Faster R-CNN are compared on the Wider Face database. Section \ref{subsec:exp_wider} evaluates the proposed MS-FRCNN against other recently published face detection methods on the Wider Face database.
% In Section \ref{subsec:exp_fddb}, our MS-FRCNN is also evaluated on the challenging FDDB face database. Finally, we analyze some cases when MS-FRCNN fails in detecting human faces.

\subsection{Experiments on WIDER FACE Dataset}
\label{subsec:expWider}
\subsubsection*{Data description}
WIDER FACE is a public face detection benchmark dataset. It contains 393,703 labeled human faces from 32,203 images collected based on 61 event classes from internet. The database has many human faces with a high degree of pose variation, large occlusions, low-resolutions and strong lighting conditions. The images in this database are organized and split into three subsets, i.e. training, validation and testing. Each contains 40\%, 10\% and 50\% respectively of the original databases.
The images and the ground-truth labels of the training and the validation sets are available online for experiments. However, in the testing set, only the testing images (not the ground-truth labels) are available online. All detection results are sent to the database server for evaluating and receiving the Precision-Recall curves.

In our experiments, the proposed CMS-RCNN is trained on the training set of the WIDER FACE dataset containing 159,424 annotated faces collected in 12,880 images. The trained model on this database are used in testing of all databases.

\subsubsection*{Testing and Comparison}
During the testing phase, the face images in the testing set are divided into three parts based on their detection rates on EdgeBox \cite{zitnick2014edge}. In other words, face images are divided into three levels according to the difficulties of the detection, i.e. Easy, Medium and Hard \cite{yang2016wider}. The proposed CMS-RCNN model is compared against recent strong face detection methods, i.e. Two-stage CNN \cite{yang2016wider}, Multiscale Cascade CNN \cite{yang2016wider}, Faceness \cite{yang2015faceness}, and Aggregate Channel Features (ACF) \cite{yang2014acf-multiscale}. All these methods are trained on the same training set and tested on the same testing set.

\begin{figure*}[ht]
\centering
\includegraphics[width=1.99\columnwidth]{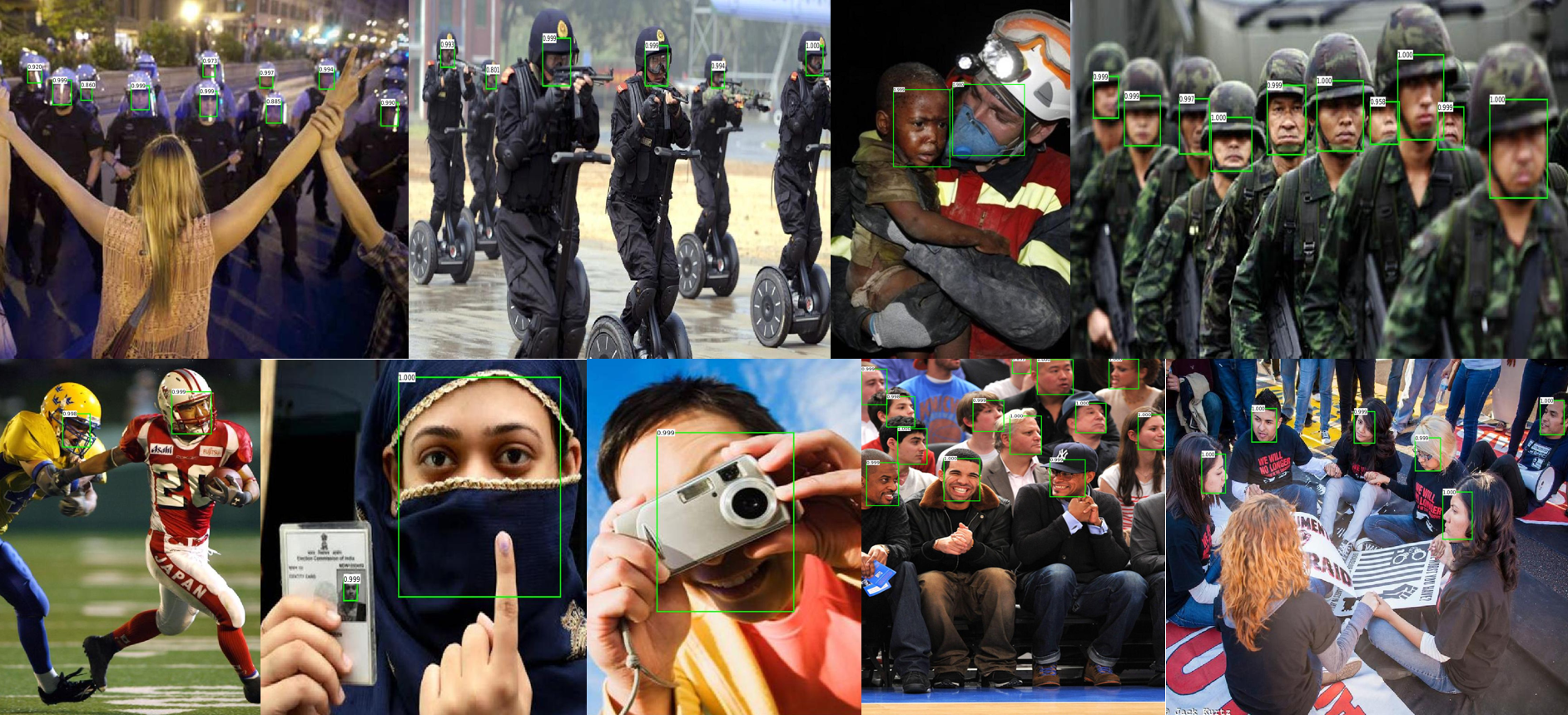}
\caption{Some examples of face detection results using our proposed CMS-RCNN method on WIDER FACE database\cite{yang2016wider}.}
\label{fig:face_wider2}
\end{figure*}

The Precision-Recall curves and AP values are shown in Figure \ref{fig:roc_wider}.
Our method outperforms those strong baselines by a large margin. It achieves the best average precision in all level faces, i.e. AP = 0.902 (Easy), 0.874 (Medium) and 0.643 (Hard), and outperforms the second best baseline by 26.0\% (Easy), 37.4\% (Medium) and 60.8\% (Hard). These results suggest that as the difficulty level goes up, CMS-RCNN can detect challenging faces better. So it has the ability to handle difficult conditions hence is more closed to human detection level. Figure \ref{fig:face_wider2} shows some examples of face detection results using the proposed CMS-RCNN on this database.

\subsubsection*{With Context v.s. Without Context}
As we show in Section~\ref{subsec:context} that human vision can benefit from additional context information for better detection and recognition, we show in this section how does explicit contextual reasoning in the network help improve the model performance.

\begin{figure}[ht]
\centering
\includegraphics[width=\columnwidth]{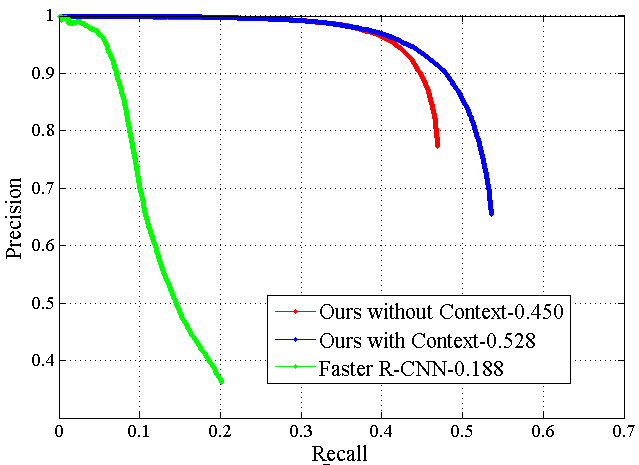}
\caption{Precision-Recall curves on the WIDER FACE validation set. The baseline (green curve) is generated by the Faster R-CNN \cite{ren2015faster} model trained on WIDER FACE training set. We show that our model without context (red curve) outperforms baseline by a large gap. With body context information, the performance gets boosted even further (blue curve). The numbers in the legend are the average precision values.}
\label{fig:context_compare}
\end{figure}

To prove this, we test our models with and without body context information on the validation set of WIDER FACE dataset. The model without body context is implemented by removing the context pipeline and only use the representation from face pipeline to compute the confidence score and the bounding box regression. We compare their performances as illustrated in Figure~\ref{fig:context_compare}. The Faster R-CNN method is setup as a baseline. 

Starting from 0 in recall, two curves of our models are overlapped at first, which means that two models perform as well as each other on some easy faces. Then the curve of model without context starts to drop quicker than the model with context, suggesting the model with context can handle the challenging conditions better when faces become more and more difficult. Thus eventually the model with context achieves a higher recall value. Additionally, the context model produces a longer PR curve, which means that contextual reasoning can help finding more faces.

\subsubsection*{Visualization of False Positives}
As it is well known that precision-recall curves get dropped due to the false positives, we are interested in the false positives produced by our CMS-RCNN model. We are curious about what object can fool our model to treat it as a face. Is it due to over-fitting, data bias, or miss labeling?

\begin{figure*}[ht]
\centering
\includegraphics[width=1.99\columnwidth]{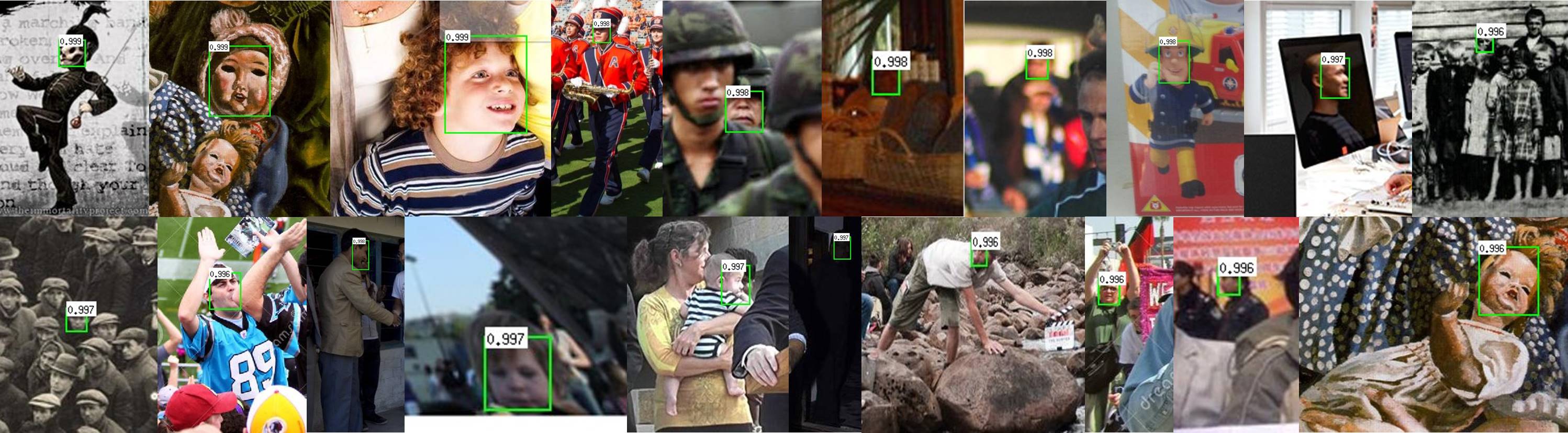}
\caption{Examples of the top 20 false positives from our CMS-RCNN model tested on the WIDER FACE validation set. In fact these false positives include many human faces not in the dataset due to mislabeling, which means that our method is robust to the noise in the data.}
\label{fig:false_pos}
\end{figure*}

In order to visualize the false positives, we test the CMS-RCNN model on the WIDER FACE validation set and pick all the false positives according to the ground truth. Then those positives are sorted by the confidence score in a descending order. We choose the top 20 false positives as illustrated in Figure~\ref{fig:false_pos} Because their confidence scores are high, they are the objects most likely to cause our model making mistakes. It turns out that most of the false positives are actually human faces caused by miss labeling, which is a problem of the dataset itself. For other false positives, we find the errors made by our model are rather reasonable. They all have the pattern of human face as well as the shape of human body.

% \begin{figure}
% \centering
% \includegraphics[width=1.0\columnwidth]{Figs/roi_compare.jpg}
% \caption{The proposed ROI candidates (upper) and the detection results (lower) using Faster R-CNN (left) and and MS-FRCNN (right). The color bar shows the confidence of each ROI region given by RPN.}
% \label{fig:msfrcnn_rois}
% \end{figure}

% \subsection{ IARPA Janus Benchmark A (IJB-A) Face Database}
% \label{subsec:iarpa}
% %http://www.nist.gov/itl/iad/ig/ijba_request.cfm

% \subsection{The Good, the Bad and the Ugly Face Database}
% \label{subsec:gbu}

\begin{figure*}[ht]
\centering 
\includegraphics[width=1.99\columnwidth]{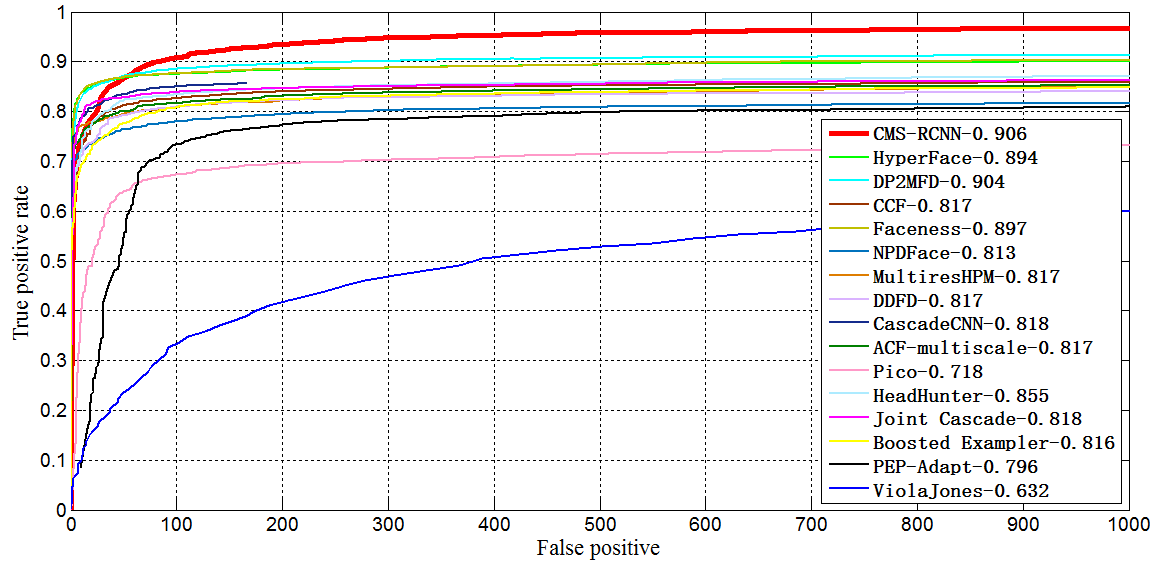}
\caption{ROC curves of our proposed CMS-RCNN and the other published methods on FDDB database \cite{fddbTech}. Our method achieves the best recall rate on this database. Numbers in the legend show the average precision scores.}
\label{fig:roc_fdd2}
\end{figure*}

\subsection{Experiments on FDDB Face Database}
\label{subsec:fddbface}
To show that our method generalizes well to other database, the proposed CMS-RCNN is also benchmarked on the FDDB database \cite{fddbTech}. It is a standard database for testing and evaluation of face detection algorithms. It contains annotations for 5,171 faces in a set of 2,845 images taken from the Faces in the Wild dataset. Most of the images in the FDDB database contain less than 3 faces that are clear or slightly occluded. The faces generally have large sizes and high resolutions compared to WIDER FACE.
We use the same model trained on WIDER FACE training set presented in Section \ref{subsec:expWider} to perform the evaluation on the FDDB database.

\begin{figure*}[ht]
\centering
\includegraphics[width=1.99\columnwidth]{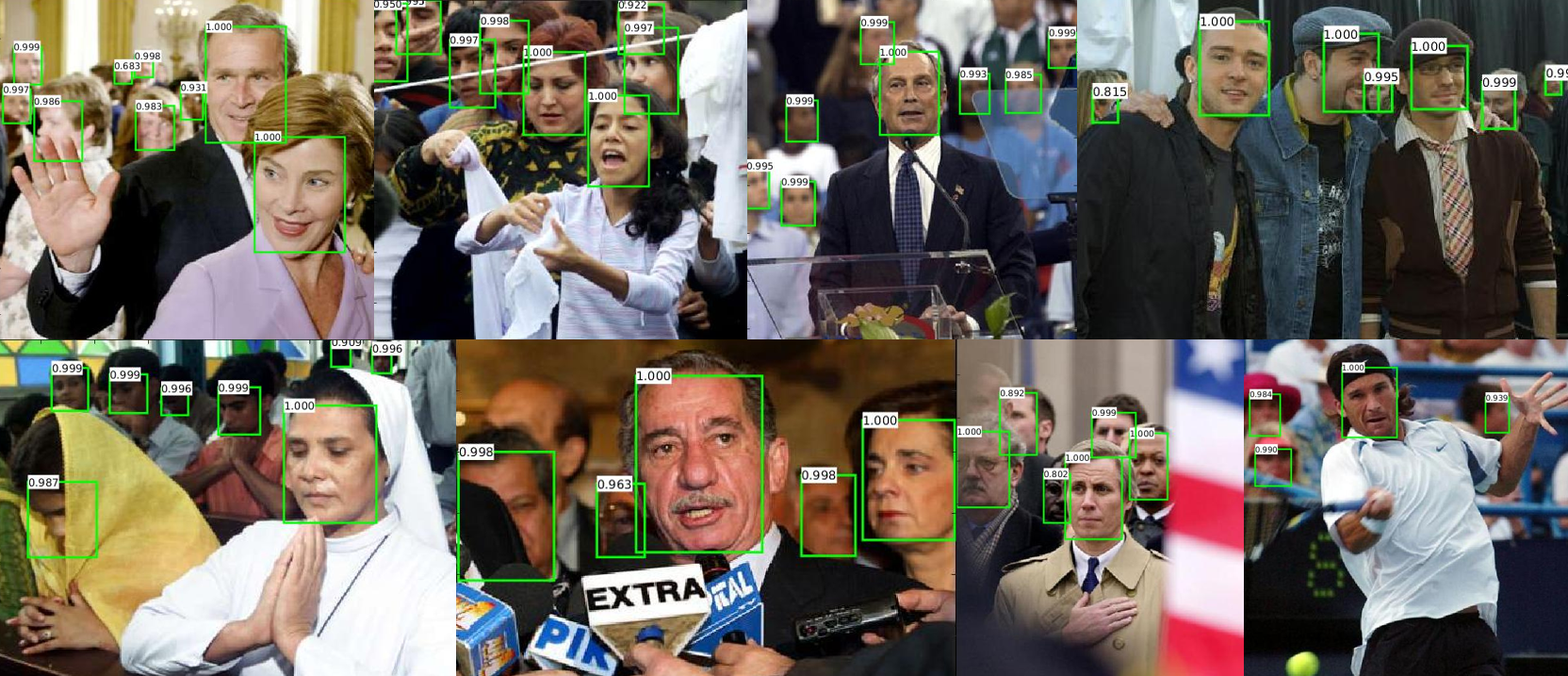}
\caption{Some examples of face detection results using our proposed CMS-RCNN method on FDDB database \cite{fddbTech}.}
\label{fig:face_fdd2}
\end{figure*}

The evaluation is performed based on the discrete criterion following the same rules in PASCAL VOC Challenge \cite{everingham2010pascal}, i.e. if the ratio of the intersection of a detected region with an annotated face region is greater than 0.5, it is considered as a true positive detection. The evaluation is proceeded following the FDDB evaluation protocol and compared against the published methods provided in the protocol, i.e. HyperFace \cite{ranjan2016hyperface}, DP2MFD \cite{ranjan2015dp2mfd}, CCF \cite{yang2015ccf}, Faceness \cite{yang2015faceness}, NPDFace \cite{liao2014npdface}, MultiresHPM \cite{ghiasi2015multireshpm}, DDFD \cite{farfade2015ddfd}, CascadeCNN \cite{li2015cascadecnn}, ACF-multiscale \cite{yang2014acf-multiscale}, Pico \cite{markuvs2013pico}, HeadHunter \cite{mathias2014face}, Joint Cascade \cite{chen2014jointcascade}, Boosted Exemplar \cite{li2014exemplar}, and PEP-Adapt \cite{li2013pep-adapt}. The proposed CMS-RCNN approach outperforms most of the published face detection methods and achieves a very high recall rate comparing against all other methods (as shown Figure~\ref{fig:roc_fdd2}). This is concrete evidence to demonstrate that CMS-RCNN robustly detects unconstrained faces. Figure~\ref{fig:face_fdd2} shows some examples of the face detection results using the proposed CMS-RCNN on the FDDB dataset.

\section{Conclusion and Future Work} \label{sec:Concl}
This paper has presented our proposed CMS-RCNN approach to robustly detect human facial regions from images collected under various challenging conditions, e.g. highly occlusions, low resolutions, facial expressions, illumination variations, etc.
The approach is benchmarked on two challenging face detection databases, i.e. the WIDER FACE Dataset and the FDDB, and compared against recent other face detection methods. The experimental results show that our proposed approach outperforms strong baselines on the WIDER FACE and consistently achieves very competitive results against state-of-the-art methods on the FDDB.

In our implementation, the proposed CMS-RCNN consists of the MS-RPN and the CMS-CNN. During training, they are merged together in an approximate joint training style for each SGD iteration, in which the derivatives w.r.t. the proposal boxes' coordinates are ignored. In the future we want to go to the fully joint training so that the network can be trained in end-to-end fashion.

\begin{figure*}[ht]
\centering
\includegraphics[width=1.99\columnwidth]{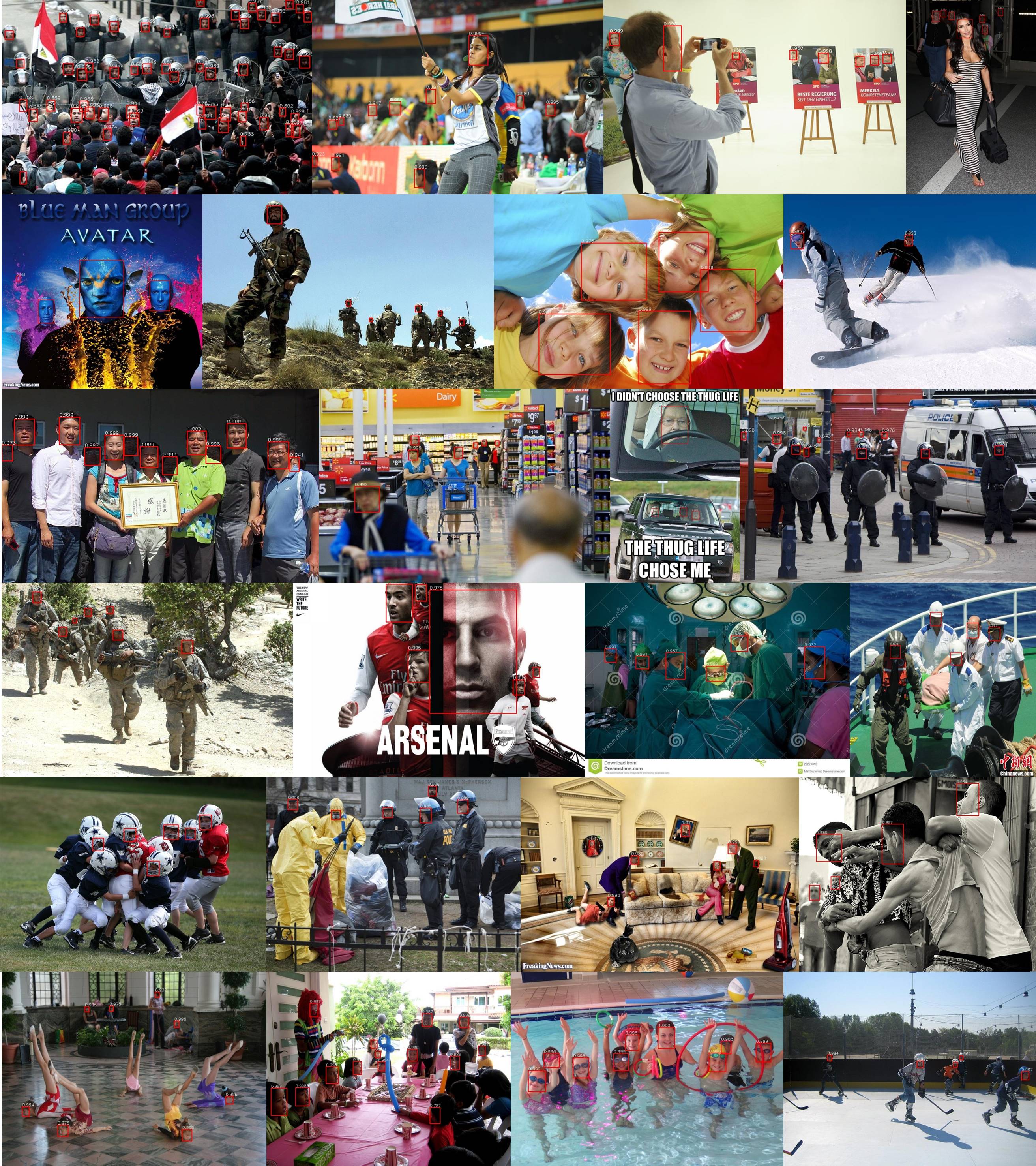}
\caption{More results of unconstrained face detection under challenging conditions using our proposed CMS-RCNN.}
\label{fig:examples}
\end{figure*}
%\section*{Acknowledgment}

%The authors would like to thank the anonymous reviewers for their insightful comments and invaluable suggestions. This research was supported under grant W911NF-09-1-0273 from the Army Research Office and by cooperative agreement W911NF-10-2-0028 from the Army Research Laboratory. The authors appreciate all members of the CMU Biometrics Center for numerous valuable discussions. In addition, we would especially like to thank Dr. Sung Won Park for allowing us to use her implementation resources for experimental comparisons. We also thank Liu et al. \cite{Liu2013} for making their method implementation available online so that we can compare to.

\ifCLASSOPTIONcaptionsoff
  \newpage
\fi



%\bibliographystyle{IEEEtran}

%\begin{thebibliography}{1}
%\bibitem{c1}
%J. Galbally, J. Fierrez, J. Ortega-Garcia, Vulnerabilities in Biometric Systems: Attacks and Recent Advances in Liveness Detection, in \emph{Proc. Spanish Workshop on Biometrics}, 2007.
%\end{thebibliography}

{\small
\bibliographystyle{IEEEtran}
\bibliography{Refs}
}

\end{document}